\documentclass[letterpaper, 10 pt, conference]{ieeeconf}  

\IEEEoverridecommandlockouts                              

\usepackage{graphicx}
\usepackage{array,booktabs}
\usepackage{hyperref}
\usepackage[x11names]{xcolor}
\usepackage[noadjust]{cite}

\newcommand{\viewOne}[1]{\leavevmode\color{DodgerBlue1}{#1}}
\newcommand{\viewTwo}[1]{\leavevmode\color{IndianRed2}{#1}}
\newcommand{\viewThree}[1]{\leavevmode\color{Chartreuse4}{#1}}

\usepackage{algorithm}
\usepackage{algpseudocode}
\usepackage{algorithmicx}

\algrenewcommand{\algorithmiccomment}[1]{{\scriptsize{\textcolor{blue} {\ttfamily// #1}}}}

\usepackage{listings}

\title{\LARGE \bf
Efficient symbolic planning with views
}

\author{Stephan Hasler$^{1}$, Daniel Tanneberg$^{1}$, and Michael Gienger$^{1}$
\thanks{$^{1}$Honda Research Institute Europe GmbH, Germany
        {\tt\small \{stephan.hasler, daniel.tanneberg, michael.gienger\}@honda-ri.de}}%
}

\begin{document}

\maketitle
\thispagestyle{empty}
\pagestyle{empty}


\begin{abstract}
Robotic planning systems model spatial relations in detail as these are needed for manipulation tasks. In contrast to this, other physical attributes of objects and the effect of devices are usually oversimplified and expressed by abstract compound attributes. This limits the ability of planners to find alternative solutions. We propose to break these compound attributes down into a shared set of \emph{elementary attributes}. This strongly facilitates generalization between different tasks and environments and thus helps to find innovative solutions. On the down-side, this generalization comes with an increased complexity of the solution space. Therefore, as the main contribution of the paper, we propose a method that splits the planning problem into a sequence of views, where in each view only an increasing subset of attributes is considered. We show that this \emph{view-based} strategy offers a good compromise between planning speed and quality of the found plan, and discuss its general applicability and limitations.
\end{abstract}


\section{INTRODUCTION} \label{section:introduction}
Robotic planning systems in literature often model manipulation related aspects in detail while more high-level aspects, e.g., the working principles of devices in the environment, are strongly oversimplified. So, e.g., the effect of kitchen devices is represented with attributes like \emph{cooked} or \emph{toasted}.
With such compound attributes the system cannot generalize across devices, because their meaning and the relations between them are very abstract and unclear. As a consequence, compound attributes can only be used in a specific narrow domain.

We propose the use of a more general, physical attribute space, called \emph{elementary attribute} space, where compound attributes are broken down into inseparable, \textit{atomic}, concepts, like temperature, color, and state of matter. This yields a domain-agnostic representation that strongly facilitates the transfer between different tasks and environments. Being able to solve more tasks and transfer knowledge between them increases the usefulness of a robotic system.

However, by unifying many environments into a single domain, the domain gets more complex. Therefore, optimal planning might be very slow and satisficing planning might produce very long and implausible plans. To circumvent this, we propose to split the attribute space hierarchically, and solve a sequence of simpler planning problems instead. This is possible because the attributes of our general domain can be split into different groups between which a certain independence can be assumed.
\begin{figure}[thpb]
  \centering
  \includegraphics[width=1.0\columnwidth]{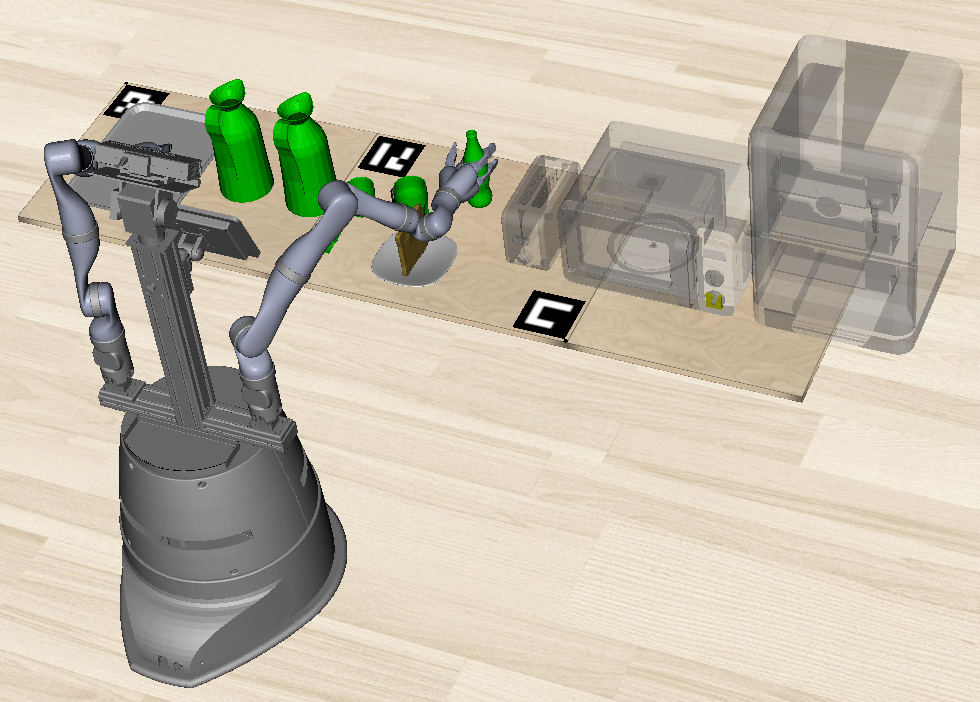}
  \caption{Snapshot of the simulation environment. 
  For clarity, only some symbolically modelled objects and devices are omitted. 
  The robot can move in one dimension along the table and manipulate objects with its two arms.}
  \label{figure:table_scene}
  \vspace{-5pt}
\end{figure}
E.g, for making iced tea, an agent can plan to make water hot and cool it down afterwards, without considering in which container the used water should be or who will manipulate the devices. We call the use of a subset of attributes a \emph{view} and the sequential approach \emph{view-based planning} (VBP). The view-based planning is the main contribution of the paper, while we will use an elementary attribute based domain to showcase the effectiveness of the method. The generated plans are executed in simulation currently, a snapshot of which is shown in Fig.~\ref{figure:table_scene}.

We are interested in getting rather short and reasonable plans quickly because this is crucial for usage in larger interactive systems~\cite{gienger2018human,muhlig2020knowledge} for cooperative intelligence~\cite{sendhoff2020cooperative}. In such systems the planning goal can be inferred from text~\cite{devlin2018bert,brown2020language,tellex2020robots,thoppilan2022lamda,deigmoeller2022situational}, and after planning human-machine-interaction (HMI) can be used to communicate the plan~\footnote{A video of the planner integrated in an HMI setting can be seen here: \url{https://youtu.be/U6HoZAy0fwo}}~\cite{wang2023explainable} and get immediate feedback. Such feedback and demonstrations by a human can be the input for learning and refining planning representations~\cite{konidaris2018from,kim2020learning,rodriguez2021learning,silver2021learning,tanneberg2023learning}.

We present related work in Sec.~\ref{section:related_work} and outline the view-based planning approach in Sec.~\ref{section:method}. In Sec.~\ref{section:problem} we describe the planning domain using elementary attributes that we use for the evaluation of the efficiency of the VBP in Sec.~\ref{section:results}. Finally, we discuss further aspects and limitations in Sec.~\ref{section:limitations} and close with the conclusion in Sec.~\ref{section:conclusion}.


\section{RELATED WORK} \label{section:related_work}
The idea of the proposed view-based planning shares commonalities with hierarchical planning~\cite{bercher2019survey} and hierarchical task networks (HTN)~\cite{georgievski2015htn}.
The key idea to both approaches is to solve planning problems leveraging hierarchical knowledge or structure.
However, there are differences in which kind of hierarchy knowledge is required and how it is used.
HTN planning uses hierarchical decompositions of (abstract) actions into sequences of actions, until a totally ordered sequence of primitive actions -- executable actions -- can be extracted (in HTN planning it is more common to talk in tasks instead of actions).
Classical planning is concerned with finding a sequence of actions to transfer an initial state into one or several goal states. 
HTN planning in contrast refines an initial task into a sequence of primitive, i.e., executable, subtasks~\cite{ghallab2004automated}. The decompositions need to be known, and encode additional domain knowledge that is usually costly designed by experts.

There is also a controversy due to this requirement of well-conceived and well-structured domain knowledge, encoding more of the solution~\cite{georgievski2015htn}.
VBP also requires additional knowledge, which is the definition of the views. However, this information is easier to provide and, importantly, independent of any task definition.
By using the elementary attributes that transfer across tasks and environments, the view definitions can also be transferred, and selecting subsets of predicates is faster and easier than specifying task decompositions.
Solving HTN planning requires more complex approaches~\cite{bercher2019survey}, and designing good (domain-independent) heuristics is difficult~\cite{holler2019guiding}.
Contrary, VBP is more like a meta planning approach, that can be instantiated with different black-box and state of the art planning algorithms, and, hence, can use established methods and heuristics.

Another approach with similarities to VBP is task and motion planning (TAMP) in robotics~\cite{garrett2021integrated}.
In TAMP, task planning is used to find a high-level plan of symbolic actions, that serve as a skeleton to guide and, hence, simplify, the motion planning for long horizon tasks.
While TAMP provides a sequence of higher-level actions that is usually fixed and sequentially executed (and solved) by low-level controllers, each view in VBP provides a plan with guidance from a previous view, but is not restricted to the previous action order.
According to the view definition, actions get partially grounded~\cite{gnad2019learning}, providing more guidance to the next view.
This partial grounding is similar to the least commitment principle~\cite{weld1994introduction} in partial-order planners~\cite{nguyen2001reviving}.


\section{VIEW-BASED PLANNING} \label{section:method}
We consider classical symbolic planning problems, given as $T=(\mathcal{A},s,g)$, where $\mathcal{A}$ is the set of actions, $s$ the initial state, and $g$ the goal. We refer to the set of predicates that are used in $T$ as $\mathcal{P}$. For view-based planning, we assume a definition $V=(\mathcal{P}^1, \mathcal{P}^2, ..., \mathcal{P}^N)$, with $N$ being the number of views, is given that splits the original problem into a sequence of simpler problems by considering a smaller subset of predicates first and by adding the remaining predicates in later steps, such that $\mathcal{P}^{n}\supset \mathcal{P}^{n-1}$ and $\mathcal{P}^{N}=\mathcal{P}$.

The general algorithm of view-based planning is shown in Alg.~\ref{algorithm:planning}. 
Each iteration starts with generating a view by applying the function \texttt{FILTER} on the problem.
\texttt{FILTER} keeps only predicates that are in the given set $\mathcal{P}^n$, and is applied to
the initial state, the goal, and the preconditions and effects of each action. 
As a consequence, a parameter of an action can be become obsolete and is removed, or an action can lose all effects and is removed in whole.
Next, the function \texttt{SOLVE} tries to find the sequence $\Pi^n$ of actions that solves the filtered problem. \texttt{SOLVE} can be implemented by any symbolic planner. Please note, even when using an optimal planner as \texttt{SOLVE} the overall VBP approach is just satisficing, as it does not solve the problem at once.  
For the first view, the described process starts with the original set of actions $\mathcal{A}^{'}=\mathcal{A}$. 
For subsequent views, $\mathcal{A}^{'}$ is the result of the function \texttt{MODIFY} that additionally depends on the solution of the previous view and the set of all actions $\mathcal{A}_{C}$ given to \texttt{SOLVE} so far. \texttt{MODIFY} has two steps. In the first step, each action in $\Pi^n$ is used to (partially) ground the corresponding action in $\mathcal{A}$. Thus, an action in $\mathcal{A}$ can potentially produce several (partially) grounded actions. All of these are added to $\mathcal{A}^{'}$. In the second step, \texttt{MODIFY} adds all actions from $\mathcal{A}$ to $\mathcal{A}^{'}$ that are not in $\mathcal{A}_{C}$, i.e., all actions that have been filtered out in whole in the previous views. With these steps, \texttt{MODIFY} ensures that actions survive that are either used in a plan or did not get a chance yet, while actions die out that \texttt{SOLVE} decided against. Finally, the plan for the last view $\Pi^N$ is also a solution of the original problem $T$, because all predicates are used, $\mathcal{P}^{N}=\mathcal{P}$, and thus $g^N=g$, $s^N=s$, and no preconditions or effects are missing in the actions $\mathcal{A}^N$.
\begin{algorithm}[!t]
    \caption{View-based planning}
    \label{algorithm:planning}
    \begin{algorithmic}[1] 
        \Require Problem $T=(\mathcal{A},s,g)$, views $V=(\mathcal{P}^1, \mathcal{P}^2, ..., \mathcal{P}^N)$
        \Ensure Plan for the final view $\Pi^N$
        \Statex \Comment{Set for remembering considered actions.}
        \State $\mathcal{A}_{C} \gets \{\}$
        \Statex \Comment{Reference for modified original actions.}
        \State $\mathcal{A}^{'} \gets \mathcal{A}$
        \For{$n \in [1..N]$}
        \Statex \hspace{1em} \Comment{Simplify problem by keeping predicate subset.}
        \State $(\mathcal{A}^n,s^n,g^n) \gets$ \Call{Filter}{($\mathcal{A}^{'},s,g), \mathcal{P}^n$}
        \Statex \hspace{1em} \Comment{Solve problem with standard planner.}
        \State $\Pi^n \gets$ \Call{Solve}{$(\mathcal{A}^n,s^n,g^n)$}
        \Statex \hspace{1em} \Comment{Update considered actions.}
        \State $\mathcal{A}_{C} \gets \mathcal{A}_{C} \cup \mathcal{A}^n$
        \Statex \hspace{1em} \Comment{Modify original actions with current solution.}
        \State $\mathcal{A}^{'} \gets$ \Call{Modify}{$\mathcal{A}, \Pi^{n}, \mathcal{A}_{C}$}
        \EndFor
    \end{algorithmic}
\end{algorithm}

View-based planning balances the complexity of the planning problem in each view. Over time, \texttt{FILTER} keeps more predicates, which makes it more difficult, while \texttt{MODIFY} grounds or removes more and more actions, which simplifies the problem. Solving $N$ simpler problems can be much faster than solving one difficult problem, because planning time often depends exponentially on problem complexity. This speedup can also enable to use an optimal planner for a sub-problem instead of a satisficing one, in this way balancing plan quality against planning time. We will evaluate this in Sec.~\ref{section:results} after describing our example domain using elementary attributes in Sec.~\ref{section:problem}.


\section{EXAMPLE DOMAIN} \label{section:problem}
For a proof of concept we have constructed an example planning domain in which the effects of devices are modelled by means of elementary attributes. We envision that a qualitatively similar domain could later be learned with the interactive system. 

Attributes are described by a set of discrete symbolic values, e.g., $temperature \in \{cold, warm, hot\}$. We will refer to these values as predicates. The list of currently modelled predicates is given in Tab.~\ref{table:predicates}.
\begin{table}[thpb]
\caption{The modelled predicates.}
\label{table:predicates}
\begin{center}
\begin{tabular}{lll}
\toprule
\textbf{Predicates} & \textbf{Attribute} & \textbf{Group}\\
\midrule
\{absent, present\} & presence &  $\mathcal{R}$ - required\\
\{hot, warm, cold\} & temperature & $\mathcal{E}$ - elementary\\
\{solid, liquid, gas\} & state of matter & $\mathcal{E}$ - elementary\\
\{gray, purple, ... , no\_color\} & color & $\mathcal{E}$ - elementary\\
\{chamomile, mint, ... , no\_aroma\} & aroma & $\mathcal{E}$ - elementary\\
\{salt, bitter, sweet, sour, no\_taste\} & taste & $\mathcal{E}$ - elementary\\
\{light, dark\} & light & $\mathcal{E}$ - elementary\\
\{dry, wet\} & wetness & $\mathcal{E}$ - elementary\\
\{soft, hard\} & hardness & $\mathcal{E}$ - elementary\\
\{clean, dirty\} & cleanness & $\mathcal{E}$ - elementary\\
\{whole, granular\} & granularity & $\mathcal{E}$ - elementary\\
\{inside, outside\} & content & $\mathcal{S}$ - spatial\\
\{close, distant\} & distance & $\mathcal{S}$ - spatial\\
\{permeable, impermeable\} & permeability & $\mathcal{S}$ - spatial\\
\{closed, open\} & openness & $\mathcal{D}$ - device\\
\{off, on\} & activity & $\mathcal{D}$ - device\\
\bottomrule
\end{tabular}
\end{center}
\end{table}
The predicates are assigned to groups that are later used to define the views $V$. The group labelled \emph{elementary} represents the elementary attributes that should allow generalization across devices. The used devices and objects in our example are:
\begin{itemize}
    \item \textbf{Devices}: \textit{toaster, microwave, blender, hair\_dryer, fridge, contact\_grill}
    \item \textbf{Objects}: \textit{water, water\_pitcher, milk, milk\_pitcher, cola, cola\_bottle, white\_bread, plate, pringles, pringles\_box, glass, mug, tray, chamomile\_tea\_bag, mint\_tea\_bag, table, robot\_hand}
\end{itemize}
For each device there is an action that defines under which conditions the device changes the elementary attributes of an object. Furthermore, some actions describe a transfer of an attribute from one object to another one, e.g., aroma going from a \textit{tea\_bag} into \textit{water}. Besides this, there are several manipulation related actions, like \textit{put\_in, get\_out, approach\_with, pour, open, close, switch\_on, switch\_off}. 
In Tab.~\ref{table:actions}, an example action is shown for each of the three mentioned categories.

\begin{table*}[t]
\caption{Action descriptions in different views. 
The table shows the definition of three example actions and how these change in the three views of the plan in Fig.\ref{figure:example_plan}. A detailed explanation of this table is given at the end of Sec.~\ref{section:results}}
\label{table:actions}
\begin{center}
\begin{tabular}{ccccp{4.9cm}p{4.5cm}}
\toprule
\bf{View} & \multicolumn{3}{l}{\textbf{Parameters} (types are omitted)} & \textbf{Preconditions} (conjunction) & \textbf{Effects} (conjunction)\\
\midrule[0.5mm]
\multicolumn{6}{l}{\texttt{a) USE A MICROWAVE ?m ON AN OBJECT ?o THAT IS INSIDE A CONTAINER ?c}}\\
\midrule[0.5mm]
  & ?m & ?o & ?c & (present ?m) (wet ?o)\newline (inside ?c ?m) (inside ?o ?c)\newline (closed ?m) (on ?m) & (hot ?o) (\textbf{not} (cool ?o)) (\textbf{not} (warm ?o))\\
\midrule[0.4mm]
\viewOne{1} & \viewOne{?m} & \viewOne{?o} & $\times$ & \viewOne{(present ?m) (wet ?o)} & \viewOne{(hot ?o) (\textbf{not} (cool ?o)) (\textbf{not} (warm ?o))}\\
\cmidrule[0.1mm](l{1.5em}){1-6}
\viewTwo{2} & \viewTwo{microwave} & \viewTwo{water} & \viewTwo{?c} & \viewOne{(present ?m) (wet ?o)}\newline\viewTwo{(inside ?c ?m) (inside ?o ?c)} & \viewOne{(hot ?o) (\textbf{not} (cool ?o)) (\textbf{not} (warm ?o))}\\
\cmidrule[0.1mm](l{1.5em}){1-6}
\viewThree{3} & \viewTwo{microwave} & \viewTwo{water} & \viewThree{water\_pitcher} & \viewOne{(present ?m) (wet ?o)}\newline \viewTwo{(inside ?c ?m) (inside ?o ?c)}\newline\viewThree{(closed ?m) (on ?m)} & \viewOne{(hot ?o) (\textbf{not} (cool ?o)) (\textbf{not} (warm ?o))}\\
\midrule[0.5mm]
\multicolumn{6}{l}{\texttt{b) USE AN OBJECT ?o1 WITH MINT AROMA ON A HOT LIQUID ?o2 THAT ARE BOTH INSIDE A CONTAINER ?c}}\\
\midrule[0.5mm]
  & ?o1 & ?o2 & ?c & (mint ?o1) (hot ?o2) (liquid ?o2)\newline (inside ?o1 ?c) (inside ?o2 ?c) & (mint ?o2) (\textbf{not} (no\_aroma ?o2))\\
\midrule[0.4mm]
\viewOne{1} & \viewOne{?o1} & \viewOne{?o2} & $\times$ & \viewOne{(mint ?o1) (hot ?o2) (liquid ?o2)} & \viewOne{(mint ?o2) (\textbf{not} (no\_aroma ?o2))}\\
\cmidrule[0.1mm](l{1.5em}){1-6}
\viewTwo{2} & \viewTwo{mint\_tea\_bag} & \viewTwo{water} & \viewTwo{?c} & \viewOne{(mint ?o1) (hot ?o2) (liquid ?o2)} \newline \viewTwo{(inside ?o1 ?c) (inside ?o2 ?c)} & \viewOne{(mint ?o2) (\textbf{not} (no\_aroma ?o2))}\\
\cmidrule[0.1mm](l{1.5em}){1-6}
\viewThree{3} & \viewTwo{mint\_tea\_bag} & \viewTwo{water} & \viewThree{water\_pitcher} & \viewOne{(mint ?o1) (hot ?o2) (liquid ?o2)} \newline \viewTwo{(inside ?o1 ?c) (inside ?o2 ?c)} & \viewOne{(mint ?o2) (\textbf{not} (no\_aroma ?o2))}\\
\midrule[0.5mm]
\multicolumn{6}{l}{\texttt{c) PUT AN OBJECT ?o INSIDE A CONTAINER ?c FROM A HAND ?h}}\\
\midrule[0.5mm]
  & ?o & ?c & ?h & (solid ?o) (inside ?o ?h) (open ?c) & (inside ?o ?c) (\textbf{not} (inside ?o ?h))\\
\midrule[0.4mm]
\viewOne{1} & \viewOne{?o} & $\times$ & $\times$ & \viewOne{(solid ?o)} & $\times$\\
\cmidrule[0.1mm](l{1.5em}){1-6}
\viewTwo{2} & \viewOne{?o} & \viewTwo{?c} & \viewTwo{?h} & \viewOne{(solid ?o)} \viewTwo{(inside ?o ?h)} & \viewTwo{(inside ?o ?c) (\textbf{not} (inside ?o ?h))}\\
\cmidrule[0.1mm](l{1.5em}){1-6}
\viewThree{3} & \viewThree{mint\_tea\_bag} & \viewThree{water\_pitcher} & \viewThree{robot\_hand} & \viewOne{(solid ?o)} \viewTwo{(inside ?o ?h)} \viewThree{(open ?c)} & \viewTwo{(inside ?o ?c) (\textbf{not} (inside ?o ?h))}\\
  & \viewThree{glass} & \viewThree{tray} & \viewThree{robot\_hand} & \viewOne{(solid ?o)} \viewTwo{(inside ?o ?h)} \viewThree{(open ?c)} & \viewTwo{(inside ?o ?c) (\textbf{not} (inside ?o ?h))}\\
  & \viewThree{water\_pitcher} & \viewThree{microwave} & \viewThree{robot\_hand} & \viewOne{(solid ?o)} \viewTwo{(inside ?o ?h)} \viewThree{(open ?c)} & \viewTwo{(inside ?o ?c) (\textbf{not} (inside ?o ?h))}\\
  & \viewThree{water\_pitcher} & \viewThree{fridge} & \viewThree{robot\_hand} & \viewOne{(solid ?o)} \viewTwo{(inside ?o ?h)} \viewThree{(open ?c)} & \viewTwo{(inside ?o ?c) (\textbf{not} (inside ?o ?h))}\\
\bottomrule
\end{tabular}
\end{center}
\end{table*}

The actions that change \emph{spatial} predicates $\mathcal{S}$ are mostly independent of $\emph{elementary}$ predicates $\mathcal{E}$, and the actions that change \emph{device}-specific predicates $\mathcal{D}$ are fully independent of \emph{spatial} and \emph{elementary} ones. Based on this, we defined the views $V$ to be:
\begin{itemize}
    \item $\mathcal{P}^1=\mathcal{R} \cup \mathcal{E}$ \hspace{0.5em} (elementary predicates first)
    \item $\mathcal{P}^2=\mathcal{R} \cup \mathcal{E} \cup \mathcal{S}$ \hspace{0.5em} (add spatial relations)
    \item $\mathcal{P}^3=\mathcal{R} \cup \mathcal{E} \cup \mathcal{S} \cup \mathcal{D}$ \hspace{0.5em} (add simple device states)
\end{itemize}
The \emph{required} predicates $\mathcal{R}$ are used in all views to prevent the \texttt{FILTER} function in Alg.~\ref{algorithm:planning} from dropping essential action parameters, e.g., \emph{a\_microwave} must always be a parameter of the \emph{use\_microwave} action, otherwise the planner could produce the effect of a microwave without having one.


\section{RESULTS} \label{section:results}
In this section we start with a quantitative evaluation of the performance gain using view-based planning. 
Afterwards, we discuss the qualitative effect of using it in combination with the elementary attributes. 
A fair quantitative comparison to HTNs is hardly possible, as they use a strongly different problem representation.

In the upper part of Tab.~\ref{table:times} we list the example goals used for the evaluation. We also combine pairs of these simpler goals to get more difficult problems. Please note, that the combined goal 0+1 is not solvable as the \emph{cola} cannot be \emph{hot} and \emph{cold} at the same time.
\begin{table*}[thpb]
\caption{Example planning goals (upper), and the corresponding planning times and costs (lower).}
\label{table:times}
\begin{center}
\begin{tabular}{ccc|cc|cc|cc|cc|cc}
\toprule
\bf{ID} & \bf{Description}\\
\midrule
0 & \multicolumn{12}{l}{cola, hot, inside glass, inside tray}\\
1 & \multicolumn{12}{l}{cola, cold, inside glass, inside tray}\\
2 & \multicolumn{12}{l}{milk, hot, inside mug, inside tray}\\
3 & \multicolumn{12}{l}{water, mint, cold, inside glass, inside tray}\\
4 & \multicolumn{12}{l}{white\_bread, hot, hard, brown, inside plate, inside tray}\\
5 & \multicolumn{12}{l}{white\_bread hot, hard, brown, salty, inside plate, inside tray}\\
\midrule
    &  \multicolumn{4}{c|}{\bf{Downward without VBP}} &   \multicolumn{8}{c}{\bf{Downward with VBP}} \\
    &   \multicolumn{2}{c}{S} &   \multicolumn{2}{c|}{O} &  \multicolumn{2}{c}{SSS} &  \multicolumn{2}{c}{OOO} &  \multicolumn{2}{c}{OSO} &  \multicolumn{2}{c}{SSO} \\
    & Time & Cost & Time & Cost & Time & Cost & Time & Cost & Time & Cost & Time & Cost  \\
\midrule
 4  &   0.304 & 8    &   0.319 & 8    &   0.664 & 8    &   0.663 & 8    &   0.661 & 8    &   0.662 & 8  \\
 1  &   0.306 & 10   &   0.881 & 10   &   0.660 & 10   &   0.664 & 10   &   0.658 & 10   &   0.660 & 10 \\
 2  &   0.307 & 15   &   3.111 & 11   &   0.660 & 12   &   0.661 & 11   &   0.658 & 11   &   0.656 & 11 \\
 0  &   0.310 & 15   &   2.840 & 11   &   0.658 & 12   &   0.658 & 11   &   0.658 & 11   &   0.657 & 11 \\
 5  &   0.305 & 15   &   0.737 & 15   &   0.674 & 15   &   0.683 & 15   &   0.673 & 15   &   0.674 & 15 \\
 3  &   0.308 & 26   &  20.294 & nan  &   0.700 & 26   &   1.379 & 20   &   0.707 & 20   &   0.710 & 20 \\
1 + 5  &   0.311 & 26   &  20.294 & nan  &   0.691 & 26   &   4.811 & 25   &   0.891 & 25   &   0.893 & 25 \\
0 + 2  &   0.308 & 28   &  20.295 & nan  &   0.679 & 20   &   1.754 & 19   &   0.678 & 19   &   0.677 & 19 \\
1 + 2  &   0.306 & 28   &  20.296 & nan  &   0.672 & 23   &  18.458 & 22   &   0.703 & 22   &   0.702 & 22 \\
2 + 5  &   0.314 & 30   &  20.299 & nan  &   0.684 & 27   &   3.705 & 26   &   0.827 & 26   &   0.827 & 26 \\
0 + 5  &   0.309 & 30   &  20.296 & nan  &   0.690 & 27   &   4.908 & 26   &   0.899 & 26   &   0.900 & 26 \\
1 + 3  &   0.309 & 35   &  20.298 & nan  &   0.718 & 32   &  20.476 & nan  &   0.797 & 26   &   0.798 & 26 \\
2 + 3  &   0.314 & 36   &  20.302 & nan  &   0.718 & 37   &  20.477 & nan  &   0.828 & 28   &   0.829 & 28 \\
0 + 3  &   0.310 & 39   &  20.296 & nan  &   0.714 & 35   &  20.472 & nan  &   0.786 & 26   &   0.785 & 26 \\
3 + 5  &   0.316 & 41   &  20.293 & nan  &   0.732 & 41   &  20.487 & nan  &   5.835 & 35   &   5.772 & 35 \\
0 + 1  &  20.336 & nan  &  20.294 & nan  &   0.224 & inf  &   0.223 & inf  &   0.224 & inf  &   0.225 & inf\\
\bottomrule
\end{tabular}
\end{center}
\end{table*}
In the lower part of Tab.~\ref{table:times} we report the required times and computed costs for different planners. 
As baselines we used Fast Downward with alias 'lama-first'~\cite{richter2010lama} as a satisficing planner, referred to as S, and Fast Downward with alias 'seq-opt-lmcut'~\cite{helmert2006fast} as an optimal planner, referred to as O. The rows in Tab.~\ref{table:times} are sorted in ascending order of the cost determined by S. The reported times are the overall times including starting Fast Downward from Python and later also preparing the views. This is the fairest comparison for our use case. The actual search time of Fast Downward is considerably lower. We gave each planner a search time limit of $20$ seconds and report a cost of \textit{nan} when this time was exceeded and a cost of \textit{inf} when the planner found the problem to be unsolvable.

The satisficing planner S can solve all the goals very quickly in about 0.3 seconds. Only for the unsolvable goal 0+1 it times out. The optimal planner O times out for most goals including the unsolvable one. It finds a more optimal plan for two goals in around 3 seconds.

View-based planning invokes a standard planner per view, i.e. three times corresponding to our definition of $V$.
When opting for more optimal plans, one could use the optimal planner O three times. We refer to this approach as OOO. This variant of VBP can solve more problems than the standalone optimal planner O. It is also substantially quicker in doing so, except for the least costly goal 4. Each VBP approach and thus even OOO are satisficing planners. Still for our domain  OOO yields substantially lower costs than the standalone satisficing planner S and the same cost as the standalone optimal planner O.

When opting for even more speed, the VBP variant SSS seems a plausible choice. However, this approach is not quicker than S. This is due to the overhead to generate the views and to start Fast Downward multiple times. Interestingly, SSS often finds a less costly plan than S.

We also tested all other mixtures of planners \{SSO, SOS, SOO, OSS, OSO, OOS\} which led to following insights. First, it is crucial to use a satisficing planner S for the second view, as this view deals with the complex spatial relations, for which an optimal planner often fails. And second, using a satisficing planner S for the last view usually leads to higher costs, while an optimal planner O can even throw out grounded but unnecessary actions from satisfying solutions of previous views in acceptable time. Like this, OSO and SSO offer a good compromise between speed and cost. They can solve all problems in acceptable time, while also having low costs. Furthermore, all VBP variants very quickly detect in the first view that the goal 0+1 is not solvable, while the standalone approaches S and O time out. In this way, VBP renders planning usable for interactive scenarios

\begin{figure}[thpb]
  \centering
  \includegraphics[width=0.7\columnwidth]{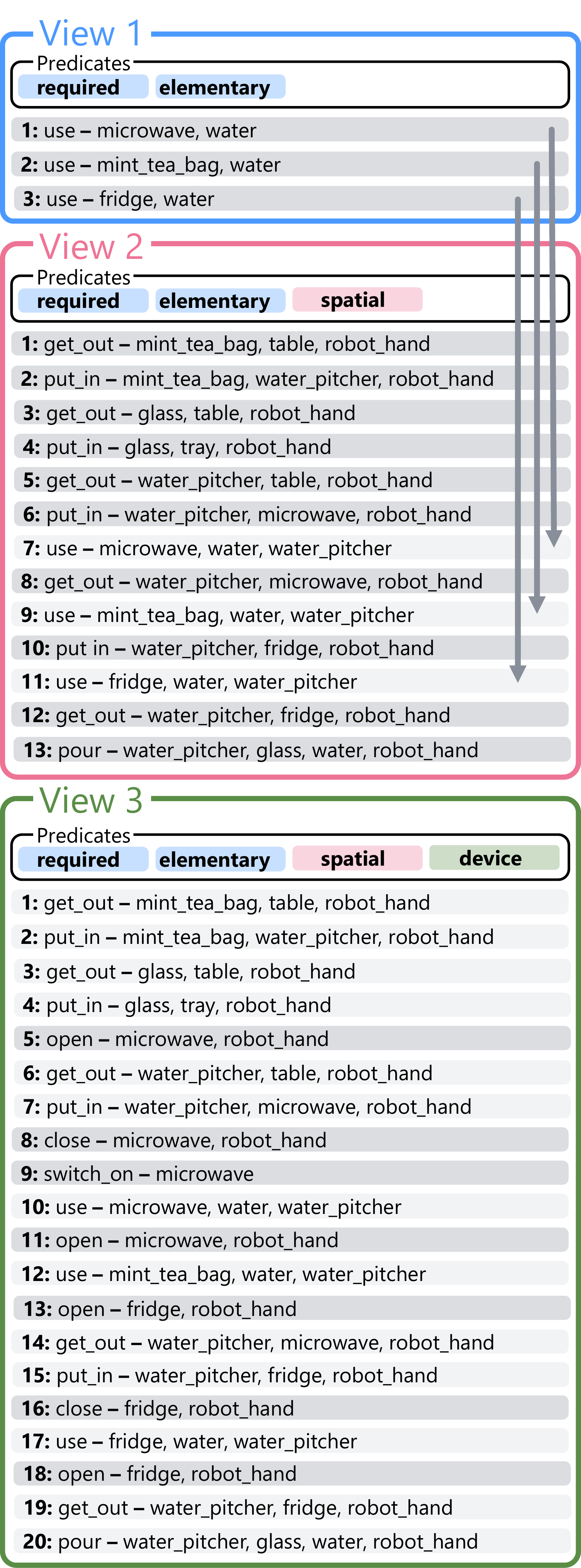}
  \caption{The view-based plan for goal 3 (mint iced tea).
    In each view the used predicate groups are shown on top, and the found plan as list of actions below. 
    Each action is given as \textit{name -- list of grounded parameters (objects)} and darker shaded actions are introduced in the corresponding view.
  }
  \label{figure:example_plan}
  \vspace{-15pt}
\end{figure}

We will discuss some qualitative aspects of VBP on example of the most difficult single goal 3, which stands for making mint iced tea. The view-based solution for this goal is shown in Fig.~\ref{figure:example_plan}. 
In view 1, the \emph{water} is made \emph{hot} with the \emph{microwave}, then the \emph{tea\_bag} makes it \emph{minty}, finally the \emph{fridge} makes it \emph{cold}. Please note, that this does not mean that the \emph{mint\_tea\_bag} must be put into the \emph{water} after using the \emph{microwave}, but instead it is the first thing the planner does in views 2 and 3, often. 
This is an advantage of view-based planning over HTNs. It does not elaborate on sequential sub-tasks, instead it deals with the full temporal complexity of the problem in each view and thus can give a more cost efficient temporal plan. In subsequent views, we see how the plan gets more detailed and incorporates more predicates and actions until a plan is found in the full representation. 

However, view-based planning splits the problem semantically. Therefore, in general it is also prone to find less cost efficient plans compared to a standard planner that sees the full problem at once. We observed that e.g. for goal 2 (making hot milk). The standard planners always use the \emph{hair\_dryer} for this as it requires only few manipulation actions (\emph{get\_out, switch\_on, approach\_with}), while view-based planning could use the \emph{microwave} instead. This requires more manipulation actions (\emph{open, get\_out, put\_in, close, switch\_on}), but in view 1 this is not visible to the planner and in later views the planner cannot revert its decision to use the \emph{microwave}. Using the \emph{hair\_dryer} in exchange of the \emph{microwave} shows that modelling the world as a single domain with elementary features helps to generalize across devices and to find innovative solutions.

In Tab.~\ref{table:actions} we highlight how some example actions change during the iteration of views while solving goal 3, where each view is highlighted by a color, and the parameters and predicates are colored according to the view they were introduced or specialized. 
This should provide more insight into the workings of Alg.~\ref{algorithm:planning}. In view 1, the function \texttt{FILTER} removes some preconditions and effects of all shown actions. As as consequence action a) and b) lose the obsolete parameter (\emph{?c - a\_container}). Action c) has no effects after filtering and is removed completely. Action a) and b) are used in the plan $\Pi^1$. Therefore, as preparation for view 2, the function \texttt{MODIFY} grounds both partially. \texttt{FILTER} in view 2 keeps the \emph{inside} predicate, and, as a result, the parameter (\emph{?c - a\_container}) survives this time for all actions. The plan of view $\Pi^2$ uses all actions. Therefore, \texttt{MODIFY} grounds the last parameter (\emph{?c - a\_container}) for action a) and b). It also fully grounds action c) four times, this means for each different occurrence in $\Pi^2$. Finally, \texttt{FILTER} keeps the remaining predicates \emph{\{open, closed, on\}} for view 3.


\section{LIMITATIONS AND DISCUSSION} \label{section:limitations}
As mentioned before, view-based planning requires a certain independence between groups of predicates.
Otherwise, it might not be efficient or even incomplete, i.e. not be able to find an existing solution. Currently, the predicate space, including the elementary attributes, is designed in a suitable way to fulfill this condition. Nevertheless, following our initial motivation, we think that many different environments and tasks are covered by the proposed general domain, and, therefore, also view-based planning will be applicable. In general, the construction of views given the problem must be investigated in more theory in coming research and might also lead to a method that can validate or even determine $V$ algorithmically.

A more general way to make VBP complete would be to allow re-planning on earlier views in case no plan could found on a later view.
Currently this is not possible, so e.g. the planner might decide in view 1 to use a \emph{microwave} to heat something, but in a later view the robot might miss the skill to \emph{open} it. With re-planning VBP would in general be similar to TAMP~\cite{garrett2021integrated}. In addition, the search of an alternative plan can also be guided by a human, or the human can execute a missing skill or teach it to the robot.

Finding innovative solutions using elementary attributes is more costly and might also not be desirable in some situations. 
We plan to extend the overall approach in a way that common solutions, as, e.g., demonstrated by a human, are preferred and proposed first. Only when the common solution is not applicable, the use of the elementary attribute space and view-based planning should be considered on demand to find an alternative solution. 

View-based planning very quickly finds a high-level plan in view 1, and later refines it in the following views that usually take more time to be solved. This shows some relation to anytime planners (see e.g.~\cite{likhachev2008anytime,richter2010lama,ponzoni2019ample}). 
The difference is that the high-level plan has a different abstraction level and is not directly executable by the robot. 
Nevertheless, it is a complete abstract plan, based on which interaction with a human can already take place.


\section{CONCLUSION} \label{section:conclusion}
We proposed the use of an elementary attribute space to unify different more specific domains and thus allow stronger generalization, e.g., as shown among the effects of different house hold devices. 
We showed that this generalization comes with an increased complexity of the search space which prevents standard planners to find good solutions in reasonable time. 
To counteract this, we proposed view-based planning that exploits the independence of some predicates to split the planning problem into a sequence of simpler problems. 
We showed that view-based planning can produce low cost plans in reasonably short time even for complex goals, where a standard optimal planner breaks down. 
This makes it especially suitable for interactive planning scenarios in human-machine-interaction setups.
Although, the used representation is hand designed, we argue that the overall approach is general, as many tasks and environments are covered by the elementary attribute space and thus view-based planning will be applicable likewise.


\addtolength{\textheight}{-0cm}


\bibliographystyle{IEEETran.bst}
\bibliography{references}


\end{document}